\documentclass[conference]{IEEEtran}
\ifCLASSINFOpdf
\else
\fi

\usepackage{amsmath,amsfonts}
\usepackage{hyperref}
\usepackage{array}
\usepackage{textcomp}
\usepackage{url}
\usepackage{cite}
\usepackage{multirow}
\usepackage[linesnumbered,ruled,vlined]{algorithm2e}
\usepackage{booktabs}
\usepackage{graphicx}
\usepackage{threeparttable}
\usepackage[font=scriptsize,labelfont={sf,bf},textfont=sf,labelsep=period,skip=2pt]{caption}
\usepackage{tabularx}
\newcolumntype{R}[1]{>{\raggedleft\arraybackslash}p{#1}}
\newcolumntype{L}[1]{>{\raggedright\arraybackslash}p{#1}}
\newcolumntype{C}[1]{>{\centering\arraybackslash}p{#1}}
\newcolumntype{Y}{>{\raggedright\arraybackslash}X}
\newcolumntype{Z}{>{\raggedleft\arraybackslash}X}
\newcolumntype{A}{>{\centering\arraybackslash}X}
\usepackage{tikz}
\usepackage[caption=false,font=scriptsize,labelfont=sf,textfont=sf]{subfig}
\definecolor{cBindOk}{HTML}{118438}
\definecolor{cBindMid}{HTML}{CC8008}
\definecolor{cRigidFlow}{HTML}{991A19}
\definecolor{cTorsoRigid}{HTML}{2C509A}
\definecolor{cNonRigidRes}{HTML}{247942}
\usetikzlibrary{shapes, arrows, positioning, fit, backgrounds, calc, decorations.pathmorphing}
\tikzstyle{startstop} = [rectangle, rounded corners, minimum width=3cm, minimum height=1cm, text centered, draw=black, fill=red!30]
\tikzstyle{io} = [trapezium, trapezium left angle=70, trapezium right angle=110, minimum width=3cm, minimum height=1cm, text centered, draw=black, fill=blue!30]
\tikzstyle{process} = [rectangle, minimum width=3cm, minimum height=1cm, text centered, draw=black, fill=orange!30]
\tikzstyle{arrow} = [thick,->,>=stealth]
\begin{document}
%
\title{MeshBind: Point-to-Mesh Binding for Non-Rigid Scene Flow Automatic Annotation}
%
%
%

\author{Yinqi~Chen$^{1,*}$, E~Li$^{1}$, Shuaijun~Wang$^{1}$, Weiyuan~Peng$^{1}$, Meiying~Zhang$^{1}$, and Qi~Hao$^{1}$%
\thanks{This work is jointly supported by the National Natural Science Foundation of China (62261160654), the Shenzhen Fundamental Research Program (JCYJ20220818103006012), and the Shenzhen Major Project for Science and Technology (KJZD20231023092600001).}%
\thanks{$^{1}$Yinqi Chen, E Li, Shuaijun Wang, Weiyuan Peng, Meiying Zhang, and Qi Hao are with the Research Institute of Trustworthy Autonomous Systems, Southern University of Science and Technology, Shenzhen 518055, China.}%
\thanks{$^{*}$Corresponding author: Yinqi Chen.}%
}
%
%

\markboth{Chen \MakeLowercase{\textit{et al.}}: MeshBind: Point-to-Mesh Binding for Non-Rigid Scene Flow Automatic Annotation}
{Chen \MakeLowercase{\textit{et al.}}: MeshBind: Point-to-Mesh Binding for Non-Rigid Scene Flow Automatic Annotation}

%



\maketitle
\begin{abstract}
  Scene flow estimation is essential for many autonomous driving (AD) and robotic applications, but obtaining ground-truth scene flow for non-rigid objects such as pedestrians is highly difficult. We present \emph{MeshBind}, an automatic annotation method that uses mesh models to generate per-point scene flow for non-rigid objects. The novelty of this work is threefold: 1) developing a \emph{multi-view module} that reconstructs an articulated mesh model from six surround cameras and registers it to the LiDAR frame; 2) developing a \emph{local-frame binding technique} that attaches each point to the mesh via a per-triangle tangent-normal frame, invariant under rigid motion yet adaptive to articulated deformation; 3) developing a \emph{three-layer flow registration pipeline} that combines ego-vehicle motion compensation, box binding, and mesh binding to produce full-scene per-point flow. We implement MeshBind in \emph{SF-Studio}, an interactive annotation tool that visualises flow and allows human mesh corrections with automatic recomputation. We also release \emph{SFB} (SUScape-Flow Benchmark), a large-scale benchmark built on the public SUScape dataset\footnote{https://suscape.net/}, covering 41k frames with per-point non-rigid pedestrian annotations and a Rigid/Semi-Rigid/Non-Rigid evaluation protocol. Experiments show that fine-tuning on SFB reduces three-way endpoint error (EPE) by 2--3$\times$, and brings non-rigid EPE to the same level as rigid EPE.
\end{abstract}


%
\IEEEpeerreviewmaketitle

\section{Introduction}
\label{sec:intro}

\begin{figure}[!t]
  \centering
  \begin{minipage}[t]{0.48\columnwidth}
    \centering
    \subfloat[Raw LiDAR]{%
      \includegraphics[height=1.2\columnwidth]{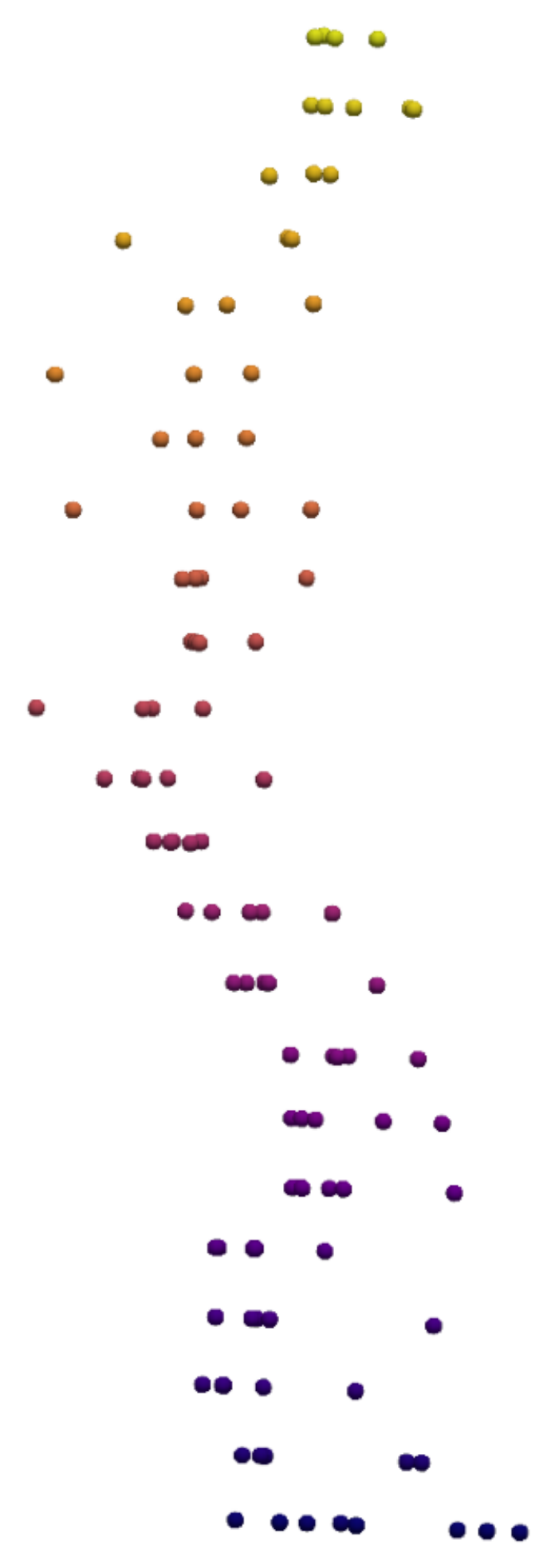}%
      \label{fig:motiv_raw}}\hfill
    \subfloat[Binding]{%
      \includegraphics[height=1.2\columnwidth]{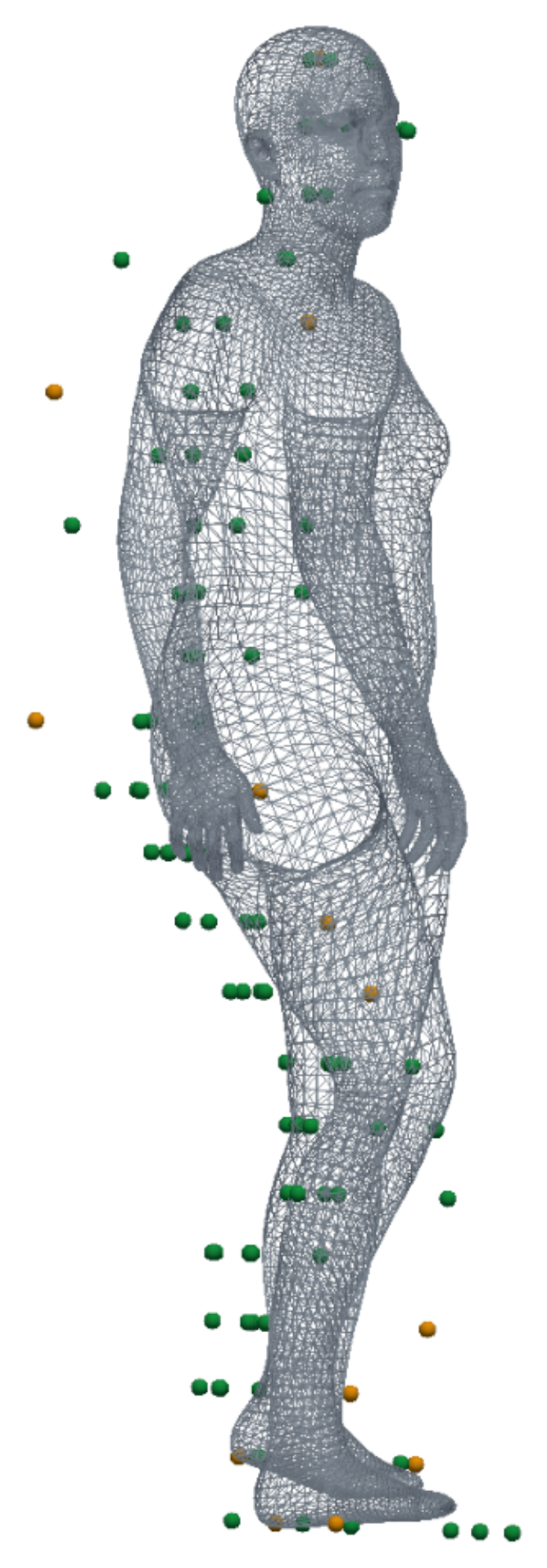}%
      \label{fig:motiv_bind}}
  \end{minipage}
  \hfill
  \begin{minipage}[t]{0.50\columnwidth}
    \centering
    \subfloat[Rigid assumption]{%
      \includegraphics[width=0.80\columnwidth]{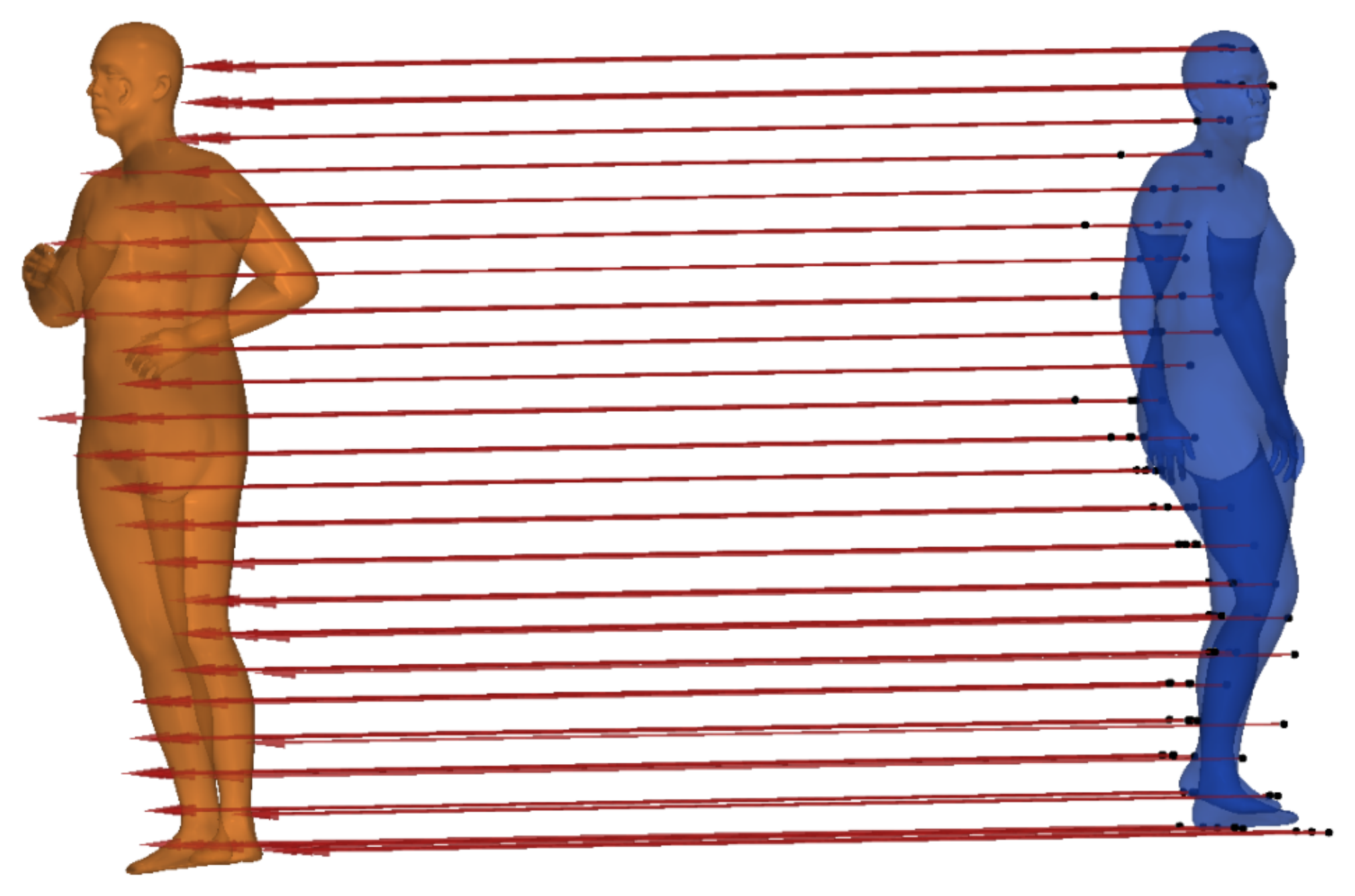}%
      \label{fig:moti_rigid}}
    \par\vspace{0.00\columnwidth}
    \subfloat[MeshBind (Ours)]{%
      \includegraphics[width=0.80\columnwidth]{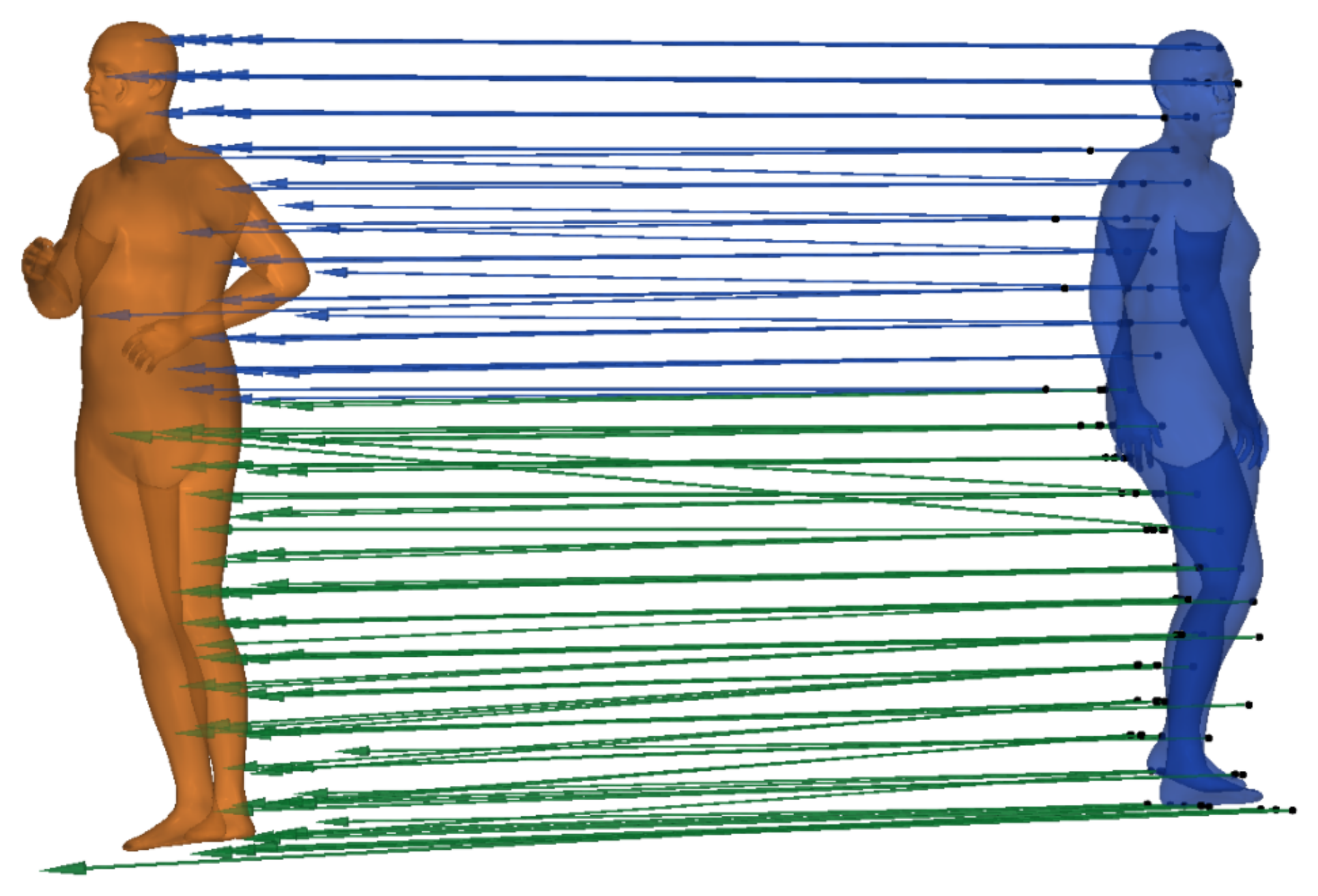}%
      \label{fig:moti_ours}}
  \end{minipage}
  \caption{Scene Flow Annotation for walking pedestrians (SUScape scene-001081, $\Delta t{=}0.5$\,s, 105 returns). (a)~Raw returns, coloured by height. (b)~Binding distance: \textcolor{cBindOk}{green} $\le 0.10$\,m / \textcolor{cBindMid}{orange} $0.10$--$0.25$\,m. (c)~The rigid convention gives every point \textcolor{cRigidFlow}{red} one translation; error concentrates at the swinging foot. (d)~MeshBind separates a \textcolor{cTorsoRigid}{blue} torso-aligned rigid part from a \textcolor{cNonRigidRes}{green} non-rigid residual at the extremities, which deviates from the rigid flow by $0.30$\,m on average (max $0.66$\,m) over a $2.13$\,m motion.}
  \label{fig:introduction}
\end{figure}

\IEEEPARstart{S}{cene} flow---the per-point 3D motion between consecutive LiDAR frames---is critical for motion segmentation~\cite{song2022ogc,chen2025semanticflow}, odometry and SLAM~\cite{zhang2014loam}, learning-based flow estimation~\cite{vedder2024zeroflow,wang2023active,zhang2024seflow}, and navigation~\cite{chen2025ssf} in AD and robotic applications.
However, most real-world benchmarks assume a single 6-DoF rigid transform per dynamic object.
This works for vehicles, but fails for walking pedestrians: leg swing, arm oscillation, and torso bending generate per-point motion that no rigid model can capture, especially at limbs (Fig.~\ref{fig:introduction}).
Waymo~\cite{sun2020scalability} acknowledges this limitation, yet justifies it by its 10\,Hz sampling rate.
For other datasets, at the lower 2\,Hz LiDAR sampling rate, the 0.5\,s articulation makes the rigid approximation qualitatively wrong.
Consequently, instance-level metrics hide systematic errors at joints, and supervision inherited from rigid assumptions structurally underrepresents true articulation.

\begin{figure*}[!t]
  \centering
  \includegraphics[width=\linewidth]{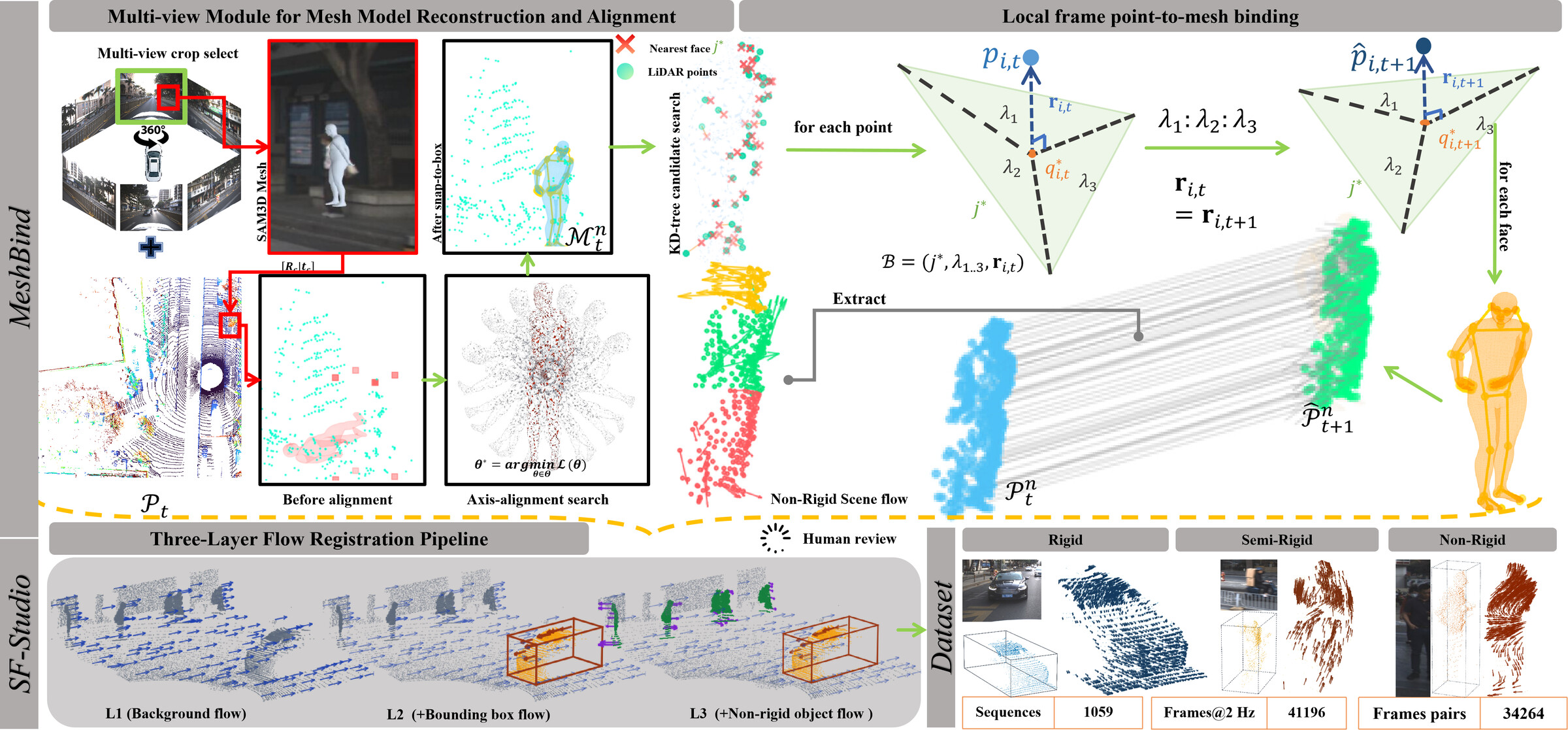}
  \caption{MeshBind annotation pipeline. \emph{Top}: the MeshBind stages---multi-view crop selection and mesh reconstruction, LiDAR-frame alignment, point-to-mesh binding with its reliability filter, and the mesh-based reconstruction that yields per-point non-rigid flow. \emph{Bottom}: the three flow layers (\emph{L1 Background}, \emph{L2 Bounding box}, \emph{L3 Non-rigid object}) produced by the method, carried in \emph{SF-Studio} for review, together with the MeshBind-annotated dataset showing one instance per motion regime.}
  \label{fig:pipeline}
  \end{figure*}

Generating per-point non-rigid flow for real LiDAR remains difficult, as manual labelling on sparse noisy point clouds is infeasible, and synthetic benchmarks~\cite{chen2024humanreg} do not replicate real sensor noise. Recent foundation models for 3D human mesh construction~\cite{yang2026sam}, built on parametric body models~\cite{loper2023smpl}, can predict detailed articulated meshes from images. Given that autonomous platforms are typically equipped with multi-view cameras, these meshes offer a continuous surface whose temporal deformation directly encodes non-rigid motion. However, several challenges need to be addressed when deploying such methods in real-world scenes:
\begin{itemize}
\item \textbf{Mesh misalignment:} The articulated meshes reconstructed from the camera follow a coordinate convention different from that of the LiDAR frame. Consequently, a direct extrinsic projection results in a persistent mismatch between the mesh and the LiDAR points; the meshes must first be registered to the LiDAR frame before any point-to-mesh binding can be established.
\item \textbf{Robust point-to-mesh binding:} A na\"ive world-coordinate offset becomes invalid as the mesh deforms. To reliably track each point through limb-level motion using a single stored tuple, the binding representation must remain invariant to rigid transformations while remaining adaptive to articulated deformations.
\item \textbf{Whole-scene flow estimation pipeline:} Scene flow must be estimated for every point in the scene, incorporating ego-motion compensation, rigid box binding, and non-rigid mesh binding. This pipeline must be designed to operate scalably and verifiably~\cite{sustechpoints2020}.
\end{itemize}

We propose \emph{MeshBind}, a mesh-guided method that annotates per-point non-rigid scene flow on real multi-beam LiDAR. The main contributions of this work include
\begin{itemize}
\item \textbf{Non-Rigid Object Mesh Model Reconstruction and Point-to-Mesh Binding Methods:} we develop a set of methods that (i)~reconstruct an articulated mesh model for each non-rigid object and register it to the LiDAR frame through a \emph{multi-view module} that exploits the six surround cameras, and (ii)~bind every LiDAR point to the mesh surface via a \emph{local-frame binding technique}---a per-triangle tangent-normal frame invariant under rigid motion yet adaptive to articulated deformation---so that mesh deformation directly drives per-point motion.
\item \textbf{Automatic Scene Flow Annotation Pipeline:} an automatic pipeline that converts multi-view images and LiDAR point clouds into large-scale, per-point non-rigid scene flow, composing ego-motion compensation, box binding, and mesh binding within a \emph{three-layer flow registration pipeline}.
\item \textbf{Interactive Annotation Tools:} \emph{SF-Studio} treats per-point non-rigid flow as a first-class label, pairing flow visualisation and per-object mesh editing with a server-backed recompute loop that re-binds an edited mesh and persists the corrected mesh binding layer.
\item \textbf{Non-Rigid Object Scene Flow Dataset and Evaluation Benchmark:} we construct \emph{SFB} (SUScape-Flow Benchmark), the first whole-scene LiDAR scene-flow benchmark with per-point non-rigid pedestrian annotations---spanning 41{,}196 frames and over 310k human detections---together with a Rigid / Semi-Rigid / Non-Rigid motion-regime evaluation protocol validated by zero-shot evaluation of eleven estimators and fine-tuned / from-scratch training of nine estimators.
\end{itemize}

\section{Related Work}
\label{sec:related}
\subsection{Scene Flow Datasets and Benchmarks}
Existing scene-flow datasets are categorized by their ground-truth generation methods: (i)~\emph{Box-based rigid} methods (e.g., Waymo~\cite{sun2020scalability}, Argoverse~2~\cite{wilson2023argoverse}, nuScenes~\cite{caesar2020nuscenes}) apply each 3D bounding box's rigid transform to all interior points, an approach that Waymo itself admits fails for non-rigid objects like pedestrians~\cite{jund2021scalable}. (ii)~\emph{CAD fitting} methods (e.g., KITTI~\cite{geiger2013vision}) use rigid 3D CAD models aligned with stereo and LiDAR data. (iii)~\emph{Synthetic} methods (e.g., Spring~\cite{mehl2023spring}, SynFlow-4k~\cite{zhang2026synflow}) render dense flow purely from simulation. (iv)~\emph{Body-model fitting} methods (e.g., HumanReg~\cite{chen2024humanreg}) fit SMPL templates to dense indoor scans, but this does not translate well to sparse automotive LiDAR.
Despite their varied approaches, none of these families provide scalable, real-world, per-point non-rigid pedestrian flow. Box-based methods, in particular, suppress limb-level articulation by enforcing rigid motion per box, causing existing estimators to miss substantial pedestrian motion on Argoverse~2~\cite{khatri2024trackflow}. The evaluation metrics are equally coarse: Argoverse-style benchmarks typically report a three-way EPE (FD/BS/FS) that blends semi-rigid human--vehicle coupling with non-rigid limb motion into a single foreground value, obscuring the distinction.
The proposed \emph{SFB} fills this gap as a new category of real-world, mesh-based non-rigid annotation. To our knowledge, it is the first whole-scene LiDAR scene-flow benchmark to offer per-point non-rigid pedestrian annotations. It also introduces a Rigid / Semi-Rigid / Non-Rigid evaluation protocol that explicitly separates motion regimes previous benchmarks cannot isolate.

\subsection{Mesh Model Reconstruction and Point-to-Mesh Binding}
Human mesh reconstruction falls into two main categories. (i)~\emph{Parametric body models} like SMPL~\cite{loper2023smpl} and SMPL-X~\cite{pavlakos2019smplx} represent the body as a differentiable mesh driven by pose and shape parameters. (ii)~\emph{Image-to-mesh regressors} use deep networks to predict these parameters from single images~\cite{bogo2016keep,kanazawa2019learning,sun2021monocular}; extensions such as HuMoR~\cite{Rempe_2021_ICCV} and WHAM~\cite{shin2024wham} add temporal and world-grounding, while foundation models like SAM~3D~Body~\cite{yang2026sam} deliver high-fidelity articulated predictions under occlusion and varying lighting.
The proposed \emph{MeshBind} builds on both approaches. Its \emph{multi-view module} runs a foundation-model regressor on the six surround images to reconstruct a watertight articulated mesh for each pedestrian. However, because this mesh follows the camera convention, a straightforward extrinsic projection does not align with the LiDAR points. Standard point-to-mesh binding can associate points with triangulated surfaces using barycentric coordinates and spatial search~\cite{ericson2005real}, but it fails to preserve a fixed offset when the surface deforms. To address this, our \emph{local-frame binding technique} stores a per-triangle local-frame residual, invariant to rigid motion yet adaptive to articulation, so that a single tuple can track each point through limb-level motion.

\subsection{Annotation Tools and Pipelines}
Annotation systems are defined by two criteria: label generation method and annotation scope. (i)~\emph{Interactive labeling} tools (LATTE~\cite{wang2019latte}, SAnE~\cite{arief2020sane}, 3D-BAT~\cite{zimmer20193dbat}, SUSTech~POINTS~\cite{sustechpoints2020}) accelerate annotation via image fusion, one-click clustering, tracking, and smart evaluation, but output cuboids or segmentation masks, not per-point motion. (ii)~\emph{Automatic pseudo-labeling} (3DSFLabelling~\cite{jiang20243dsflabelling}) generates flow pseudo-labels by optimizing per-box motion parameters.
The former is human-driven but not scalable for per-point motion annotation on sparse, noisy LiDAR; the latter is automatic yet rigid, unable to model limb-level articulation. Our developed \emph{SF-Studio} combines both: \emph{MeshBind} automatically produces per-point non-rigid flow for every pedestrian, while humans only refine the reconstructed meshes. Correcting a mesh automatically triggers re-binding and flow recomputation on the backend, enabling scalable semi-automatic annotation that remains reviewable and editable.

\section{System Setup and Problem Statement}
\label{sec:setup}

\noindent\textbf{System overview.}
As shown in Fig.~\ref{fig:pipeline}, the annotation pipeline comprises \emph{MeshBind}, the labelling method, and \emph{SF-Studio}, the interactive tool that reviews its output and feeds corrections back; the released dataset and evaluation baseline are presented in Sec.~\ref{sec:dataset}--\ref{sec:eval_protocol}.
We write $\mathcal{P}_t^n=\{p_{i,t}\in P_t:\ p_{i,t}\in B_n^t\}$ for the point cloud of pedestrian~$n$ in frame~$t$---the LiDAR points falling inside its detected box $B_n^t$---and $\hat{\mathcal{P}}_{t+1}^n=\{\hat{p}_{i,t+1}:\ p_{i,t}\in\mathcal{P}_t^n\}$ for the set of recovered positions of those points in the next frame's LiDAR frame.
MeshBind runs three modules in sequence.
(i)~\emph{Multi-view module} (Sec.~\ref{sec:multiview}) picks, per pedestrian, the largest 2D crop among the six surround-view images, reconstructs an articulated mesh model with a human-mesh foundation model, and registers it to the LiDAR frame by extrinsic calibration, a discrete axis search, and a box-centroid snap.
(ii)~\emph{Local-frame binding technique} (Sec.~\ref{sec:binding}) attaches every point $p_{i,t}\in\mathcal{P}_t^n$ to its nearest mesh triangle, storing a binding set $\mathcal{B}_t$ of 3-tuples $(j^*, \lambda_{1..3}, \mathbf{r}_{i,t})$---face index, barycentric coordinates, and local-frame residual---which is invariant to rigid motion yet follows articulated deformation; a distance cut-off rejects unreliable bindings.
(iii)~\emph{Three-layer flow registration pipeline} (Sec.~\ref{sec:decomposition}) composes ego-motion compensation, box binding, and mesh binding into whole-scene per-point flow, each layer overwriting the previous one on the points it owns.
SF-Studio (Sec.~\ref{sec:studio}) then loads the released flow for review, re-binds edited meshes, and writes the recomputed flow back, closing the annotation loop.

\noindent\textbf{Problem statement.}
We address the following problems, each of which motivates one stage of the proposed method:
\begin{enumerate}
\item \emph{How to reconstruct and align a per-pedestrian articulated mesh model} $\mathcal{M}_t^n$ from multi-view images to the LiDAR frame, such that its deformation across consecutive frames serves as a geometric proxy for non-rigid motion?
\item \emph{How to bind each point} $p_{i,t} \in \mathcal{P}_t^n$ \emph{to the reconstructed mesh} $\mathcal{M}_t^n$, yielding a tuple $(j^*, \lambda_{1..3}, \mathbf{r}_{i,t})$ that remains invariant under rigid motion yet adapts to articulation, thus enabling the point to be tracked through limb-level motion across frames?
\item \emph{How to integrate ego-vehicle motion compensation, box binding, and mesh binding} into a complete per-point flow pipeline for the whole scene, while ensuring that the process remains scalable and reviewable in practice?
\end{enumerate}


\section{Proposed Methods}
\label{sec:methods}


\subsection{Multi-view Module for Mesh Model Reconstruction and Alignment}
\label{sec:multiview}
\begin{figure}[!t]
  \centering
  \includegraphics[width=\linewidth]{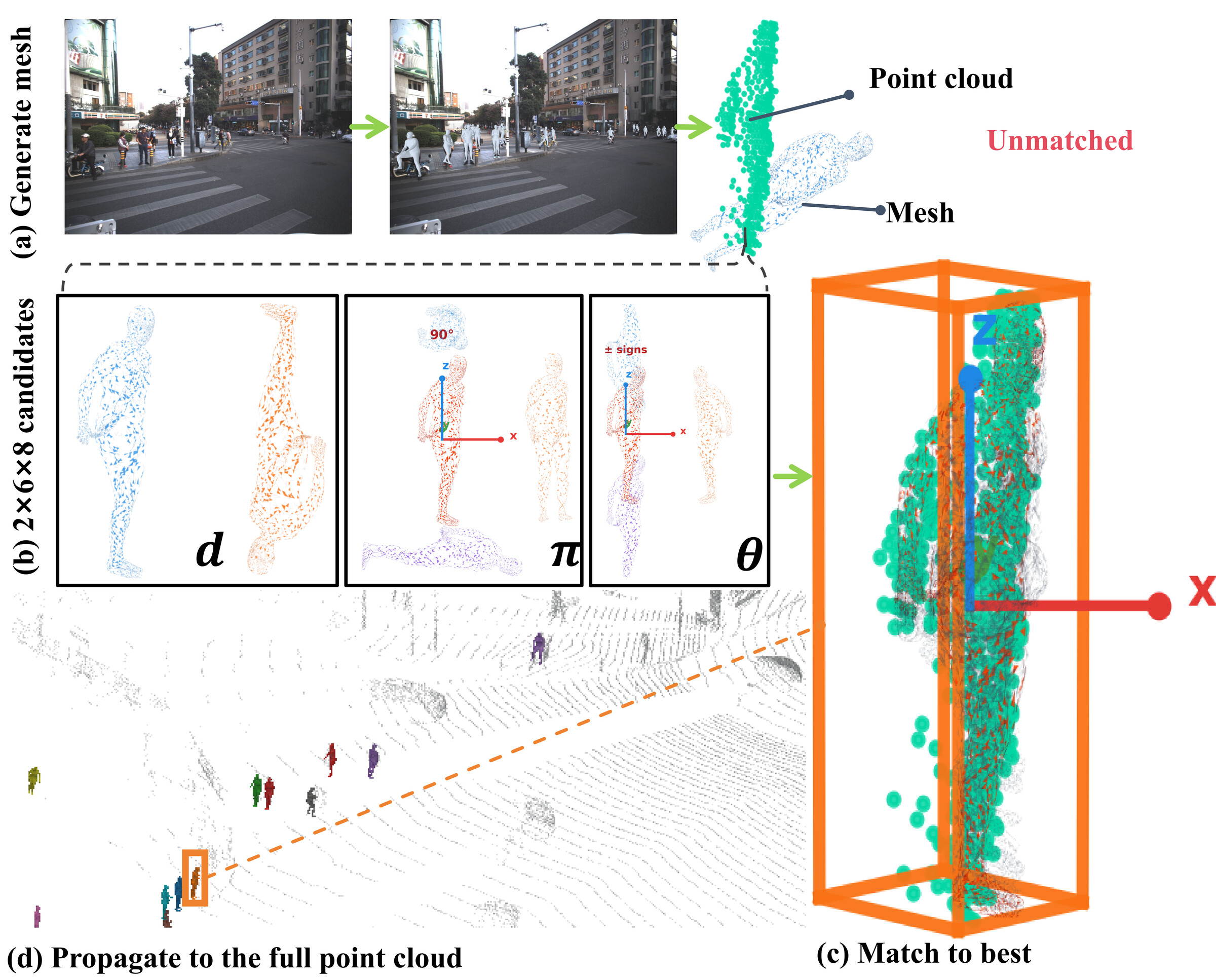}
  \caption{Multi-view module: the camera mesh is generated and unmatched with the LiDAR points (a), rotated through the 96 candidate axis corrections $\theta=(d,\pi,\sigma)$---reversing the direction $d$, permuting the axes $\pi$, flipping the signs $\sigma$ (b)---and matched to the best candidate $\theta^*$ (c), then propagated to the full point cloud (d).}
  \label{fig:meshrecover}
  \end{figure}

The reconstructed mesh model of pedestrian $n$ at frame $t$, denoted as $\mathcal{M}_t^{n}=(V_t^{n},F)$, has vertices $V_t^{n}\in\mathbb{R}^{\nu\times 3}$ and a fixed triangle list $F$.
As shown in Fig.~\ref{fig:meshrecover}, running the SAM~3D~Body~\cite{yang2026sam} foundation model on each of the six surround-view images with pedestrian detections yields a watertight articulated mesh model per detection. However, the camera convention (Y-down, Z-forward) of the foundation model may differ from the dataset's calibration convention, so the raw extrinsic projection of the mesh is unmatched with the LiDAR points, as illustrated in Fig.~\ref{fig:meshrecover}(a).
For each pedestrian $n$ visible in multiple views $\mathcal{V}_n$, we keep the prediction from the view with the largest 2D bounding-box area, $c^*_n = \arg\max_{c \in \mathcal{V}_n} A_{n,c}$, and transform it to the LiDAR frame via the extrinsic $[R_c|\mathbf{t}_c]$.
Because the extrinsic alone does not guarantee a correct registration, we search over the set $\Theta$ of $2\times6\times8{=}96$ candidate rigid corrections $\theta=(d,\pi,\sigma)$ (Fig.~\ref{fig:meshrecover}b), and we select the one $\theta^*$ that minimises the average distance from the mesh keypoints to the LiDAR points,
\begin{equation}
\mathcal{L}(\theta) = \frac{1}{NJ} \sum_{n=1}^{N} \sum_{j=1}^{J} \min_{p \in P_t} \| q_{n,j}^\theta - p \|
\label{eq:align}
\end{equation}
where $N$ is the number of pedestrians in the frame, $J$ the number of keypoints per mesh, and $q_{n,j}^\theta$ the $j$-th keypoint of pedestrian $n$ after applying transform $\theta$.
The search runs once per scene for temporal consistency.
As shown in Fig.~\ref{fig:meshrecover}(c,d), the selected $\theta^*$ aligns the mesh to the LiDAR points, and the mesh centroid is then snapped to the 3D detection box centre to correct residual foundation-model error; the same correction is propagated to the full point cloud so that every dynamic point shares the aligned reference.


\subsection{Local-Frame Binding Technique}
\label{sec:binding}
Given the aligned mesh $\mathcal{M}_t^n$ of pedestrian $n$ and its point cloud $\mathcal{P}_t^n$, we bind each $p_{i,t}\in\mathcal{P}_t^n$ to the mesh surface, storing a tuple $(j^*, \lambda_{1..3}, \mathbf{r}_{i,t})$ of face index, barycentric coordinates, and local-frame residual, which stays valid as the mesh deforms between frames. The tuple separates a \emph{topology anchor} $(j^*, \lambda_{1..3})$ (which face and where on it) from the \emph{local-frame residual} $\mathbf{r}_{i,t}$, the off-surface offset $p_{i,t}-q_{i,t}^*$ rotated into a per-triangle frame (Eq.~\ref{eq:residual}), so that it separates rigid from non-rigid motion during position recovery; its magnitude $\lVert\mathbf{r}_{i,t}\rVert$ serves as a per-point reliability measure for filtering and soft weighting.

\noindent\textbf{KD-tree candidate search.}
A KD-tree on face centroids returns the $K{=}16$ nearest faces $\mathcal{C}(p_{i,t})$ per point, excluding degenerate triangles ($\|n\|<10^{-6}$).

\noindent\textbf{Closest face and barycentric anchor.}
Among the $K{=}16$ candidates, Ericson's 7-Voronoi-region algorithm~\cite{ericson2005real} finds per face $f_j$ the closest point $q_j$ and its barycentric coordinates $(\lambda_{j,1},\lambda_{j,2},\lambda_{j,3})$; the binding face minimises the point-to-face distance,
\begin{equation}
j^* = \arg\min_{j \in \mathcal{C}(p_{i,t})} \|p_{i,t} - q_j\|, \quad
q_{i,t}^* = \textstyle\sum_{m=1}^{3} \lambda_{j^*,m}^{(t)}\, v_{j^*,m},
\label{eq:anchor}
\end{equation}
where $q_{i,t}^*{=}q_{j^*}$ is the closest point on the binding face $f_{j^*}$; the barycentric weights survive both rigid and non-rigid mesh motion.

\begin{figure}[t]
  \centering
  \includegraphics[width=\linewidth]{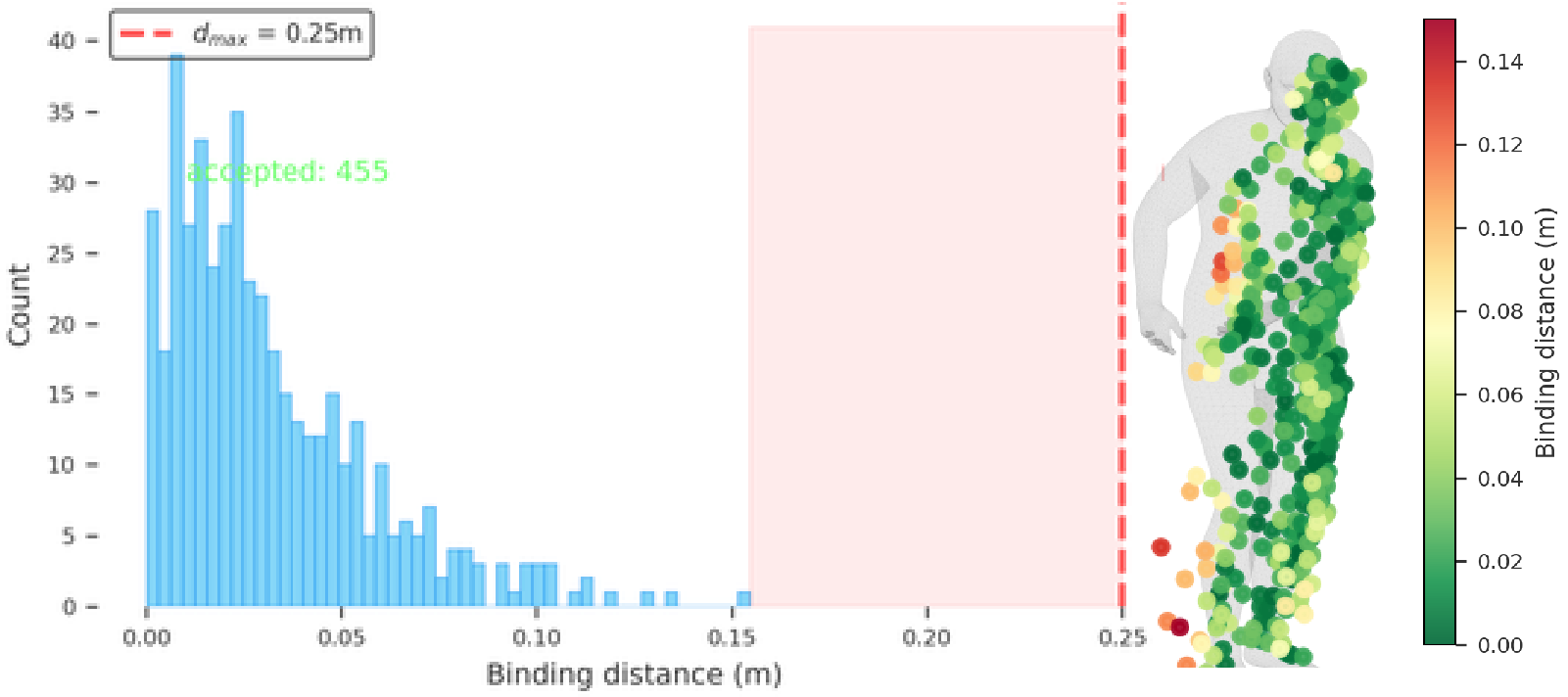}
  \caption{Binding distance on an example scene: each LiDAR point is coloured by its distance to the mesh surface, and points beyond the cut-off $d_{\max}{=}0.25$\,m are excluded from binding.}
  \label{fig:binding_filter}
  \end{figure}

\noindent\textbf{Distance filter.}
Points with binding distance $\|p_{i,t}-q_{i,t}^*\|>d_{\max}$ (default $0.25$\,m) are discarded as unreliable and fall back to box-driven flow; Fig.~\ref{fig:binding_filter} visualises this cut-off on one scene, where returns far from the mesh surface are dropped.

\noindent\textbf{Local-frame residual (core of MeshBind).}
The world-coordinate offset $\Delta_{i,t}=p_{i,t}-q_{i,t}^*$ does not rotate with the triangle, so it becomes invalid as the mesh deforms; we store it instead in a per-triangle orthonormal frame $F_j=[t_1;t_2;n]^T$ (unit, mutually orthogonal axes),
\begin{equation}
\mathbf{r}_{i,t} = F_{j^*}(p_{i,t} - q_{i,t}^*) \in \mathbb{R}^3,
\label{eq:residual}
\end{equation}
and recover the point at $t{+}1$ in the same face's deformed frame,
\begin{equation}
\hat{p}_{i,t+1} = \underbrace{\textstyle\sum_{m=1}^{3} \lambda_{j^*,m}^{(t)} v_{j^*,m}^{t+1}}_{\text{barycentric anchor}} + \underbrace{(F_{j^*}^{t+1})^T \mathbf{r}_{i,t}}_{\text{local-frame residual}}.
\label{eq:recover_b}
\end{equation}
The residual is then (i)~\emph{rigid-invariant} (unchanged under rigid triangle motion, so rigid parts such as the torso and head need no update), (ii)~\emph{deformation-adaptive} (re-projected into the deformed frame rather than drifting), and (iii)~\emph{articulated} (points on different limbs follow their own triangles, which a single 6-DoF transform cannot express).

\noindent\textbf{Reliability and soft weighting.}
Retained bindings optionally carry a soft weight $w_{i,t}=\exp(-\alpha\|\mathbf{r}_{i,t}\|)$ ($\alpha{=}10$) so reliable bindings dominate.

\begin{algorithm}[!t]
  \DontPrintSemicolon
  \SetAlgoLined
  \SetAlCapSkip{0.5em}
  \caption{Scene Flow Computation with Non-Rigid Human Replacement}
  \label{alg:flow}
  \KwIn{Source cloud $P_t$, ego-pose $(R_t,\mathbf{t}_t),\,(R_{t+1},\mathbf{t}_{t+1})$, object boxes $\{B_n\}$, binding set $\mathcal{B}_t$, target meshes $\{\mathcal{M}_{t+1}^n\}$}
  \KwOut{Per-point scene flow $\{f_i\}$}
  \BlankLine
  \For{$p_{i,t} \in P_t$}{
    $\hat{p}_{i,t+1} \gets \text{ego-compensated } p_{i,t}$ via Eq.(\ref{eq:ego})\;
  }
  \For{each matched object $B_n$}{
    $(R_n, \mathbf{t}_n) \gets \textsc{RigidUpdateFromBoxes}(B_n^{t}, B_n^{t+1})$\;
    \For{$p_{i,t} \in B_n$}{
      $\hat{p}_{i,t+1} \gets \text{box-rigid transformed } p_{i,t}$ via Eq.(\ref{eq:box})\;
    }
  }
  \For{each bound human instance $n$ with binding $\mathcal{B}_t$}{
    \For{bound point $p_{i,t}$ with tuple $(j^*, \lambda_{1..3}, \mathbf{r}_{i,t}) \in \mathcal{B}_t$}{
      $\hat{p}_{i,t+1} \gets \text{mesh binding for } p_{i,t}$ via Eq.(\ref{eq:recover_b})\;
    }
  }
  $f_i \gets \hat{p}_{i,t+1} - p_{i,t}$ \quad $\forall\, p_{i,t} \in P_t$\;
  \Return $\{f_i\}$
  \end{algorithm}

\subsection{Three-Layer Flow Registration Pipeline}
\label{sec:decomposition}
Given bindings at $t$ and target meshes at $t{+}1$, we compute per-point flow through a three-layer scheme where later layers \emph{replace} earlier predictions for affected points.

\noindent\textbf{L1: Background flow annotation.}
Transform all source points into the target frame:
\begin{equation}
\hat{p}_i^{\text{ego}} = R_{t+1}^\top (R_t p_{i,t} + \mathbf{t}_t - \mathbf{t}_{t+1}).
\label{eq:ego}
\end{equation}
Background points keep this as their final flow.

\noindent\textbf{L2: Bounding box flow annotation.}
For each detected object with matched boxes $(B_n^t, B_n^{t+1})$, whether moving or stationary, we derive a rigid transform $(R_n,\mathbf{t}_n)$ from relative box orientation and centre displacement, and replace in-box predictions:
\begin{equation}
\hat{p}_{i,t+1} \leftarrow R_n p_{i,t} + \mathbf{t}_n.
\label{eq:box}
\end{equation}
Vehicle points keep this as final; points of the human categories receive it as coarse motion before Layer~3.

\noindent\textbf{L3: Non-rigid object flow annotation.}
For bound human instances (the categories of Sec.~\ref{sec:dataset}), we replace the box prediction by the cross-frame mesh binding of Eq.(\ref{eq:recover_b}) (Fig.~\ref{fig:introduction}d): the stored topology anchor $\lambda_{j^*,m}^{(t)}$ and residual $\mathbf{r}_{i,t}$ are transported to the target mesh $\mathcal{M}_{t+1}^n$ through the same face $f_{j^*}$ (consistent topology) and its deformed local frame $F_{j^*}^{t+1}$.
Points naturally follow mesh deformation: torso points follow torso motion, upper-leg points follow thigh rotation, lower-leg points follow calf rotation.
The final flow is $f_i = \hat{p}_{i,t+1} - p_{i,t}$.
Algorithm~\ref{alg:flow} summarises the complete scene flow computation with non-rigid refinement.

\begin{figure}[!t]
\centering
\includegraphics[width=\columnwidth]{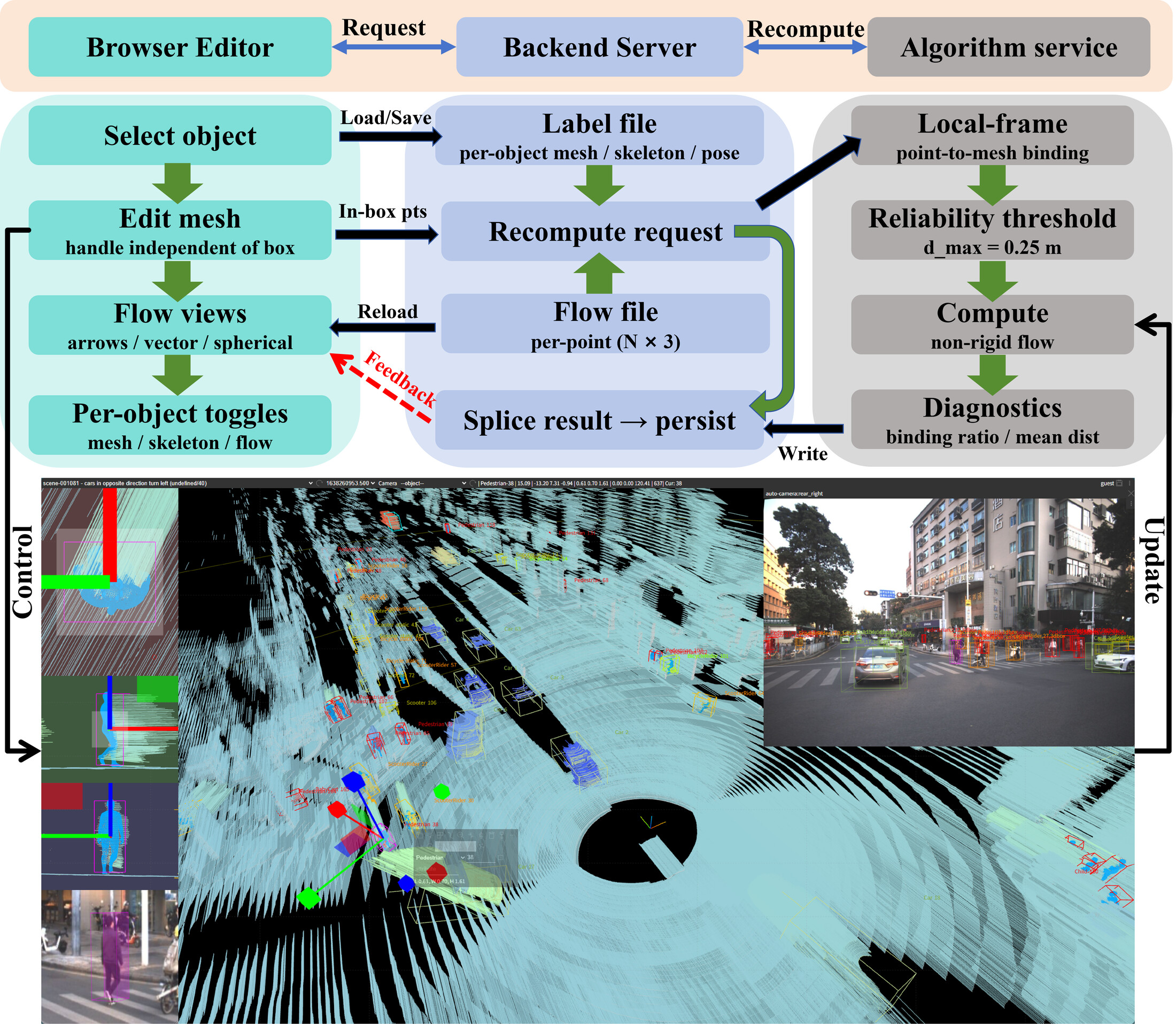}
    \caption{SF-Studio. \emph{Top}: the three-tier layout with the dashed Feedback path returning the persisted flow to the editor. \emph{Bottom}: browser-editor UI on a sample frame.}
\label{fig:studio_arch}
\end{figure}

\subsection{SF-Studio: Interactive Annotation Tools}
\label{sec:studio}
\emph{SF-Studio} extends an open-source point-cloud annotator~\cite{sustechpoints2020} with per-point flow visualisation, per-object mesh editing, and a server-backed recompute loop (Fig.~\ref{fig:studio_arch}).
It uses a three-tier layout: a \emph{Browser editor}, a \emph{Backend} owning on-disk labels (per-object mesh/skeleton/pose) and an $(N,3)$ flow file, and an \emph{Algorithm service} that runs the binding of Sec.~\ref{sec:binding} on demand.
The editor shows flow as \emph{arrows}, \emph{vector} (hue/lightness), or \emph{spherical} $(\Delta\phi,\Delta\theta,\Delta r)$, with mesh edits via a transform handle independent of the detection box.
Editing a mesh triggers recompute: the service rebinds $\mathcal{P}_t^n$ under $d_{\max}{=}0.25$\,m, returns flow plus diagnostics (binding ratio, mean binding distance), and the Backend persists the update so Feedback reloads it in the editor; low-scoring frames are queued for spot-checking. Recompute covers human categories that share the SAM~3D~Body prior; other dynamics keep box-driven flow.

\section{Experiments}

\subsection{Dataset Construction}
\label{sec:dataset}
We construct \emph{SFB} on the public SUScape multi-sensor urban driving platform, which provides synchronized LiDAR point clouds, multi-view camera images, and calibration parameters.
The benchmark comprises 1{,}059 driving sequences totalling 41{,}196 frames at 2\,Hz (train 741 / val 159 / test 159); each frame carries a 64-beam LiDAR point cloud of roughly $3.4\times10^{5}$ returns and six synchronised surround-view images.
Consecutive-frame pairs at 2\,Hz ($\Delta t{=}0.5$\,s) serve as scene-flow supervision, yielding 28{,}080 training and 6{,}026 test pairs (the 6{,}026 test pairs span 6{,}185 test frames used by the evaluation protocol below); val sequences are reserved for hyperparameter tuning.
Following Sec.~\ref{sec:studio}, the human categories sharing the SAM~3D~Body mesh prior---Pedestrian, Child, BabyCart, BicycleRider, ScooterRider, MotorcycleRider, BicycleGroup, and RoadWorker---are annotated with per-point non-rigid flow and remain editable through the SF-Studio recompute loop; all other dynamic objects, vehicles, and background returns fall back to box-driven flow.

\subsection{Evaluation Protocol and Motion-Regime Metrics}
\label{sec:eval_protocol}
We report the standard Argoverse-style three-way EPE (foreground-dynamic EPE\_FD, background-static EPE\_BS, foreground-static EPE\_FS) together with \emph{motion-regime} EPEs enabled by our per-point annotations.
Using the fine-grained category indices in each H5 frame, every valid point is assigned to one of three regimes aligned with the SUScape taxonomy: \textbf{Rigid} (vehicles: car, van, truck, bus, long vehicle, concrete truck, crane, tour car), \textbf{Semi-Rigid} (wheeled human--vehicle coupling: bicycle, bicycle-rider, bicycle-group, scooter, scooter-rider, trimotorcycle, baby-cart), and \textbf{Non-Rigid} (pedestrian, child); the regime EPE is the mean end-point error over that regime's points, exposing where rigid supervision fails.
On the test split (6{,}185 frames, $2.04{\times}10^{9}$ valid points), the three regimes account for $9.51\%$, $1.01\%$, and $0.45\%$ of returns, the rest being static background; Non-Rigid points are ${\sim}21{\times}$ scarcer than Rigid ones, motivating regime-stratified reporting.

\begin{figure}[!t]
\centering
  \includegraphics[width=\linewidth]{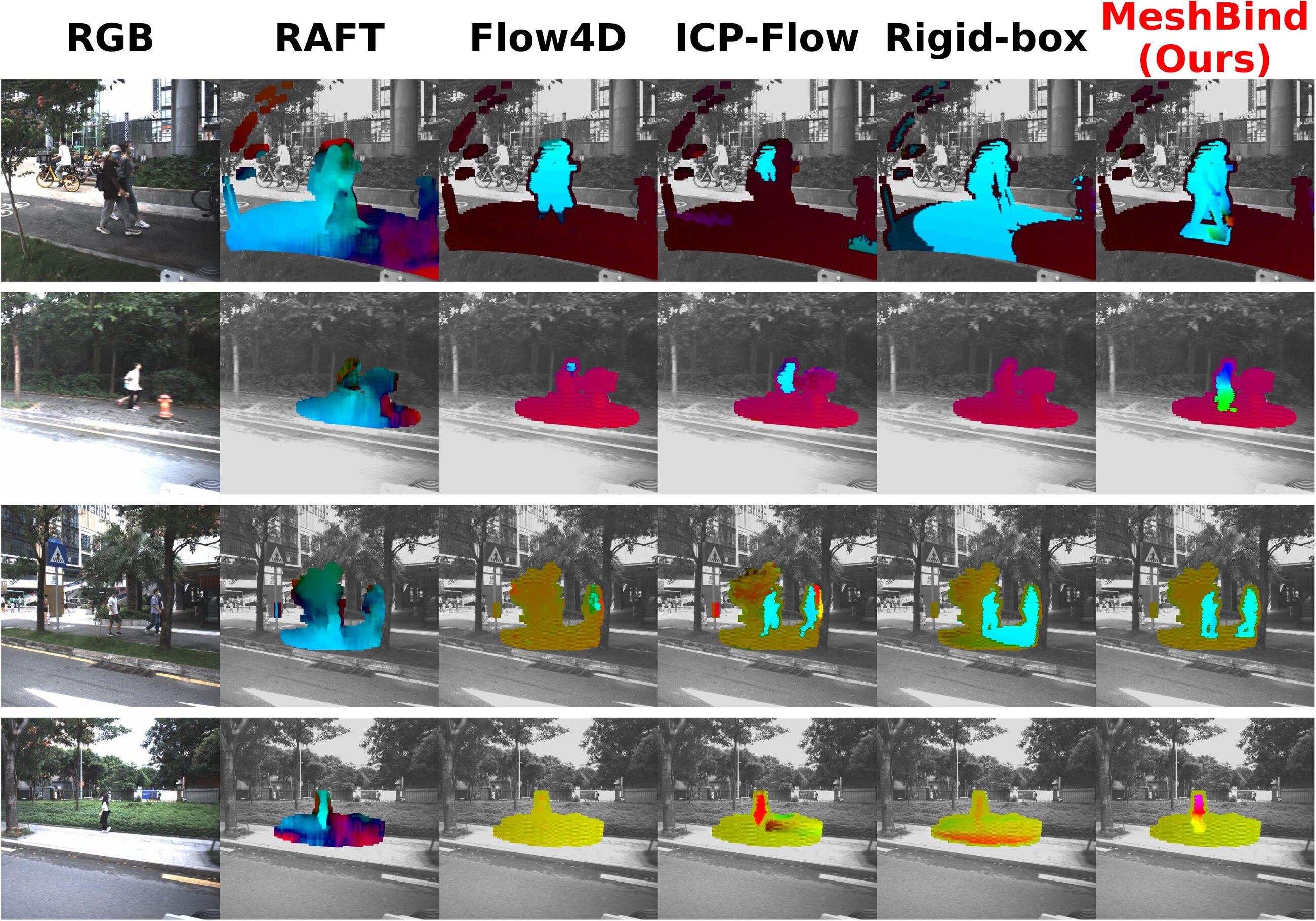}
  \caption{Cross-modal projection consistency across four scenes (front-right camera, one pedestrian per row). Columns: \emph{RGB}; \emph{RAFT} camera-motion-compensated optical flow; representative zero-shot predictions of \emph{Flow4D} (feed-forward) and \emph{ICP-Flow} (learning-free); \emph{rigid-box}; and \emph{MeshBind (Ours)} projected flow in colour (hue = direction, brightness = magnitude). MeshBind renders limb-level articulation (multi-coloured), whereas rigid-box and the baselines appear single-colour.}
  \label{fig:cross_modal}
\end{figure}
\subsection{Annotation Quality Validation}
\label{sec:quality}

\paragraph{Mesh fitting residual}
Over all 86{,}349{,}113 bound points from 30{,}510 frames across the 924 of the 1{,}059 SUScape scenes containing non-rigid or semi-rigid objects, the binding distance is sharply concentrated below $0.10$\,m (73.5\%, median $0.054$\,m, mean $0.072$\,m); the $d_{\max}{=}0.25$\,m cut-off removes only the long-range tail, retaining the remaining 26.5\% of marginal points ($0.10$--$0.25$\,m).
Across the same 30{,}510 frames, the binding success rate over 310{,}879 detected non-rigid and semi-rigid instances is high (mean $0.87$, median $0.99$), with $93.6\%$ of instances exceeding a binding ratio of $0.5$ and only $868$ ($0.3\%$) falling below $0.1$.

\paragraph{Cross-modal projection consistency}
As an independent check that does not rely on the MeshBind labels, we project 3D flow onto the image plane and compare its direction with RAFT optical flow~\cite{teed2020raft} via cosine similarity (rather than endpoint error, since the rigid-box baseline assigns zero flow to $99.95\%$ of foreground-dynamic points).
Over 300 frame pairs and 6 camera views across 15 scenes (Fig.~\ref{fig:cross_modal}), the mean cosine similarity is $0.78$ for MeshBind vs.\ $0.01$ for the rigid-box baseline, confirming that the non-rigid component of the labels is visible in 2D optical flow; the figure also shows RAFT degrading on distant or occluded pedestrians, where image-domain flow alone is insufficient and mesh-bound 3D labels provide complementary structure.

\paragraph{Synthetic data validation}
Treating SAM~3D~Body mesh vertices as GT (body-model animation, independent of MeshBind), a rigid centroid baseline yields mean EPE $0.107$\,m over 940{,}389 vertices from 56 instances, whereas local-frame recovery is exact (EPE $= 0$); on LiDAR points the gap a rigid-box label misses averages $0.41$\,m.

\begin{table}[!t]
  \centering
  \renewcommand\arraystretch{1.05}
  \caption{Zero-shot scene-flow baselines on SUScape-Flow test. Top: eight feed-forward estimators pre-trained on Argoverse~2; bottom: three learning-free baselines. EPE in metres. \emph{Fr.} = input sweeps; `---' = full-scan optimisation. Column abbreviations follow Sec.~\ref{sec:eval_protocol}: \emph{FD/BS/FS}, \emph{3-way}, \emph{Rig./Sem./NonR.} Best in bold.}
  \label{tab:benchmark_zeroshot}
  \scriptsize
  \setlength{\tabcolsep}{1.5pt}
  \begin{tabular}{lcc cccc c ccc}
    \toprule
    \textbf{Methods} & \textbf{Fr.} & \textbf{FD}$\downarrow$ & \textbf{BS}$\downarrow$ & \textbf{FS}$\downarrow$ & \textbf{3-way}$\downarrow$ & \textbf{IoU}$\uparrow$ & \textbf{Rig.}$\downarrow$ & \textbf{Sem.}$\downarrow$ & \textbf{NonR.}$\downarrow$ \\
    \midrule
    \multicolumn{10}{l}{\textit{Feed-forward (Argoverse-pretrained)}} \\
    DeFlow~\cite{zhang2024deflow}       & 2 & 2.571 & 0.0099 & 0.123 & \textbf{0.901} & 0.339 & \textbf{2.378} & 0.981 & 0.418 \\
    DeltaFlow~\cite{zhang2026deltaflow} & 2 & 2.784 & \textbf{0.0046} & 0.053 & 0.947 & 0.237 & 2.602 & 0.980 & \textbf{0.329} \\
    FastFlow3D~\cite{jund2021scalable}      & 2 & 3.498 & 0.0204 & 0.259 & 1.259 & 0.269 & 3.312 & 1.021 & 0.461 \\
    Flow4D~\cite{kim2025flow4d} (5-fr)        & 5 & 2.715 & 0.0140 & 0.153 & 0.961 & \textbf{0.459} & 2.545 & 0.965 & 0.334 \\
    SeFlow~\cite{zhang2024seflow}             & 2 & 2.580 & 0.0557 & 0.287 & 0.974 & 0.272 & 2.407 & 0.945 & 0.428 \\
    HiMo (3-fr)~\cite{zhang2025himo}         & 3 & 2.573 & 0.0526 & 0.225 & 0.950 & 0.268 & 2.399 & 0.933 & 0.419 \\
    SSF~\cite{khoche2025ssf}                  & 2 & 2.843 & 0.0159 & \textbf{0.049} & 0.969 & 0.174 & 2.645 & 1.008 & 0.396 \\
    VoteFlow~\cite{lin2025voteflow}           & 2 & \textbf{2.566} & 0.0472 & 0.266 & 0.960 & 0.295 & 2.399 & \textbf{0.931} & 0.412 \\
    \midrule
    \multicolumn{10}{l}{\textit{Learning-free (test-time optimisation)}} \\
    FastNSF~\cite{li2023fast}              & --- & \textbf{1.959} & 0.4325 & 0.927 & 1.106 & \textbf{0.210} & \textbf{1.821} & 1.086 & 0.566 \\
    NSFP~\cite{li2021nsfp}                    & --- & 2.458 & 0.1995 & 0.327 & 0.995 & 0.176 & 2.301 & 0.987 & 0.434 \\
    ICP-Flow~\cite{lin2024icp}            & --- & 2.877 & \textbf{0.0150} & \textbf{0.034} & \textbf{0.975} & 0.120 & 2.697 & \textbf{0.936} & \textbf{0.316} \\
    \bottomrule
  \end{tabular}
\end{table}

\subsection{Benchmark Baselines: Zero-Shot Scene-Flow Estimation}
\label{sec:benchmark_zeroshot}
To establish reference performance, we evaluate eleven scene-flow estimators on the test split without any SUScape fine-tuning (eight Argoverse~2-pretrained networks and three learning-free baselines; Table~\ref{tab:benchmark_zeroshot}).
Two observations stand out: the foreground-dynamic EPE\_FD (2.57--3.50\,m feed-forward, 1.96--2.88\,m learning-free) exceeds the background-static EPE\_BS by one to two orders of magnitude, confirming that dynamic objects remain the dominant error source on real driving LiDAR; and 2-frame models (DeFlow, VoteFlow) match or beat 3- and 5-frame variants (HiMo, Flow4D), the short $\Delta t{=}0.5$\,s interval rendering long temporal context less informative than on Argoverse.
The motion-regime EPEs expose an artefact of the rigid assumption: Non-Rigid is consistently the \emph{smallest} regime (avg $0.400$\,m vs.\ $2.586$\,m Rigid and $0.971$\,m Semi-Rigid, similarly in the learning-free group), not from capturing pedestrian articulation but because rigid-box supervision collapses every pedestrian point to a single 6-DoF box motion of small walking-speed displacement; combined with the scarcity of Non-Rigid points, this lets the rigid convention hide behind a small absolute error exactly where it is qualitatively wrong---ICP-Flow, enforcing a per-object rigid assumption, attains the lowest Non-Rigid EPE ($0.316$\,m) precisely where its assumption is most violated.



\begin{table*}[!t]
  \centering
  \renewcommand\arraystretch{1.15}
  \caption{Estimators trained on the SFB training split: \emph{FT} = fine-tuned from Argoverse weights, \emph{Sc} = trained from scratch (30 vs.\ 15 epochs). All EPE values in metres; $\downarrow$ lower, $\uparrow$ higher. \emph{Fr.} = input sweeps. Column abbreviations follow Sec.~\ref{sec:eval_protocol}: FD/BS/FS, 3-way EPE, foreground IoU, and the motion-regime EPEs. Best per column in bold.}
  \label{tab:downstream_flow}
  \scriptsize
  \setlength{\tabcolsep}{2.5pt}
  \begin{tabular}{lc cccccccc cccccccc}
    \toprule
    & & \multicolumn{8}{c}{\textbf{Fine-tuned (FT), 30\,ep from Argoverse weights}} & \multicolumn{8}{c}{\textbf{From-scratch (Sc), 15\,ep from random init}} \\
    \cmidrule(lr){3-10}\cmidrule(lr){11-18}
    \textbf{Methods} & \textbf{Fr.} & \textbf{FD}$\downarrow$ & \textbf{BS}$\downarrow$ & \textbf{FS}$\downarrow$ & \textbf{3-way}$\downarrow$ & \textbf{IoU}$\uparrow$ & \textbf{Rig.}$\downarrow$ & \textbf{Sem.}$\downarrow$ & \textbf{NonR.}$\downarrow$ & \textbf{FD}$\downarrow$ & \textbf{BS}$\downarrow$ & \textbf{FS}$\downarrow$ & \textbf{3-way}$\downarrow$ & \textbf{IoU}$\uparrow$ & \textbf{Rig.}$\downarrow$ & \textbf{Sem.}$\downarrow$ & \textbf{NonR.}$\downarrow$ \\
    \midrule
    DeFlow~\cite{zhang2024deflow}        & 2 & 0.203 & 0.0058 & 1.077 & 0.429 & 0.695 & 0.228 & 0.200 & 0.250 & 0.219 & 0.0062 & 1.090 & \textbf{0.439} & 0.686 & 0.242 & 0.217 & 0.255 \\
    DeltaFlow~\cite{zhang2026deltaflow}  & 2 & 0.242 & 0.0093 & 1.087 & 0.446 & 0.546 & 0.269 & 0.209 & 0.241 & 0.236 & 0.0131 & \textbf{1.076} & 0.442 & 0.441 & 0.255 & 0.214 & 0.260 \\
    FastFlow3D~\cite{jund2021scalable} & 2 & 0.267 & 0.0082 & 1.138 & 0.471 & 0.637 & 0.287 & 0.251 & 0.292 & 0.252 & 0.0078 & 1.147 & 0.469 & 0.647 & 0.271 & 0.262 & 0.315 \\
    Flow4D~\cite{kim2025flow4d}          & 5 & 0.259 & 0.0161 & 1.125 & 0.467 & 0.496 & 0.286 & 0.240 & 0.259 & 0.333 & 0.0112 & 1.082 & 0.475 & 0.578 & 0.346 & 0.279 & 0.274 \\
    SeFlow~\cite{zhang2024seflow}        & 2 & 0.201 & 0.0062 & 1.096 & 0.434 & 0.692 & 0.227 & 0.194 & 0.249 & 0.212 & 0.0063 & 1.120 & 0.446 & 0.684 & 0.236 & 0.225 & 0.258 \\
    HiMo~\cite{zhang2025himo} (3-fr)         & 3 & 0.203 & \textbf{0.0053} & \textbf{1.073} & 0.427 & \textbf{0.705} & 0.227 & 0.191 & 0.243 & 0.411 & 0.0116 & 1.112 & 0.512 & 0.626 & 0.418 & 0.324 & 0.283 \\
    SSF~\cite{khoche2025ssf}             & 2 & \textbf{0.178} & 0.0054 & 1.081 & \textbf{0.421} & 0.704 & \textbf{0.205} & \textbf{0.179} & \textbf{0.235} & \textbf{0.191} & 0.0066 & 1.127 & 0.441 & \textbf{0.699} & \textbf{0.219} & \textbf{0.188} & \textbf{0.240} \\
    TeFlow~\cite{zhang2026teflow}        & 3 & 0.280 & 0.0095 & 1.097 & 0.462 & 0.607 & 0.302 & 0.258 & 0.254 & 0.302 & 0.0176 & 1.078 & 0.466 & 0.416 & 0.319 & 0.260 & 0.263 \\
    VoteFlow~\cite{lin2025voteflow}      & 2 & 0.226 & 0.0054 & 1.136 & 0.456 & 0.691 & 0.256 & 0.209 & 0.247 & 0.244 & \textbf{0.0059} & 1.088 & 0.446 & 0.680 & 0.264 & 0.235 & 0.263 \\
    \bottomrule
  \end{tabular}
\end{table*}


\subsection{Downstream Validation: Training Estimators on SFB}
\label{sec:downstream}
The nine estimators of Table~\ref{tab:downstream_flow} are trained on the SFB training split (28{,}080 pairs) in two regimes: \emph{Fine-tuned} (FT) from Argoverse-pretrained weights for 30 epochs, and \emph{From-scratch} (Sc) from random initialisation for 15 epochs; TeFlow lacks an official Argoverse-pretrained release and therefore has no zero-shot entry in Table~\ref{tab:benchmark_zeroshot}. FT cuts the three-way EPE of every estimator with a zero-shot baseline by 2--3$\times$ (e.g.\ DeFlow $0.901 \to 0.429$, SSF $0.969 \to 0.421$).
EPE\_FS rises from $\sim$0.15\,m to $\sim$1.1\,m---a labelling consequence, not a regression: under the Argoverse protocol a box displaced below the 0.05\,m threshold is ``static,'' yet MeshBind assigns non-zero per-point flow (e.g.\ a standing pedestrian swinging their arms), which FT estimators now predict. The regime distribution inverts: Non-Rigid EPE drops from $0.400$\,m (zero-shot) to $0.252$\,m, on par with Rigid ($0.254$\,m) and above Semi-Rigid ($0.214$\,m), as estimators fit the limb-level motion MeshBind exposes; HiMo and SSF attain the lowest three-way EPE ($0.427$ / $0.421$\,m), SSF best on Non-Rigid ($0.235$\,m). Sc reaches within $+3.1\%$ of the FT mean three-way EPE ($0.460$ vs.\ $0.446$), three estimators surpassing their FT counterparts---SFB suffices as a primary training set; the Non-Rigid pattern is preserved (mean $0.268$ vs.\ Rigid $0.286$\,m) except HiMo, whose 3-frame aggregation degrades by $19.9\%$ without warm initialisation. All zero-shot-to-FT gains are significant ($p < 0.001$, paired $t$-test across 6{,}185 test frames); FT-vs-Sc differences are not, except for HiMo. The Non-Rigid gain is not a circular artefact; Sec.~\ref{sec:quality} independently validates the labels via cross-modal RAFT consistency and lossless synthetic recovery of articulation.


\subsection{Ablation Study}
\label{sec:ablation}

\paragraph{Local-frame residual vs.\ world-coordinate residual.}
We quantify the core choice of storing the residual $\mathbf{r}_{i,t}$ in a per-triangle tangent-normal frame rather than as a world-coordinate offset $\Delta_{i,t} = p_{i,t} - q_{i,t}^*$ (Sec.~\ref{sec:binding}), on 10 consecutive frame pairs from scene-001081 (80{,}373 bound points, mean triangle rotation $26.5^\circ$, $64.2\%$ of triangles rotating $> 10^\circ$).
Measuring the per-point flow gap $|f^{\text{world}}_i - f^{\text{local}}_i|$ between the two representations: on rigidly moving triangles (rotation $\le 10^\circ$) it is negligible (mean $0.005$\,m), confirming the residual's rigid invariance (the two representations agree under rigid motion, so the local frame introduces no bias), whereas on deforming triangles ($> 10^\circ$) it grows $\mathbf{6\times}$ to mean $0.031$\,m (median $0.013$\,m), confirming its deformation adaptivity (the fixed world-coordinate offset no longer follows the rotated triangle, so the two representations diverge).
Since $64\%$ of pedestrian triangles rotate beyond $10^\circ$ across $\Delta t{=}0.5$\,s, the fixed world-coordinate assumption fails for the majority of triangles, making the local frame the necessary representation for tracking articulated limb motion.

\begin{table}[!t]
  \centering
  \renewcommand\arraystretch{0.92}
  \setlength{\tabcolsep}{3.5pt}
  \abovecaptionskip=-4pt
  \caption{Residual tolerance under rigid mesh perturbation. ``No-residual'' forces $\mathbf{r}_{i,t}{=}0$; ``With-residual'' is the full MeshBind representation. Drift in metres ($\downarrow$ lower).}
  \label{tab:perturbation_ablation}
  \footnotesize
  \begin{tabular}{ccc cc}
    \toprule
    Perturbation & Coverage & \multicolumn{2}{c}{Recovery drift (m)} \\
    \cmidrule(lr){3-4}
    (m) & (\%) & No-residual & With-residual (Ours) \\
    \midrule
    0.00 & 99.9 & 0.061 & \textbf{0.000} \\
    0.02 & 67.5 & 0.146 & \textbf{0.060} \\
    0.05 & 66.8 & 0.158 & \textbf{0.063} \\
    0.08 & 63.7 & 0.156 & \textbf{0.059} \\
    0.12 & 64.1 & 0.174 & \textbf{0.068} \\
    0.15 & 61.7 & 0.186 & \textbf{0.067} \\
    0.20 & 52.8 & 0.190 & \textbf{0.074} \\
    0.25 & 38.8 & 0.210 & \textbf{0.083} \\
    \bottomrule
  \end{tabular}
  \vspace{-8pt}
\end{table}

\paragraph{Residual tolerance under mesh misalignment.}
The local-frame residual $\mathbf{r}_{i,t}$ also tolerates mesh localisation error (Table~\ref{tab:perturbation_ablation}).
Applying the \emph{same} rigid perturbation (translation up to $0.25$\,m, rotation up to $5^\circ$) to both $\mathcal{M}_t^n$ and $\mathcal{M}_{t+1}^n$, the ``No-residual'' scheme (forced $\mathbf{r}_{i,t}{=}0$) drifts $0.061$\,m even at zero perturbation---matching this scene's mean binding distance, which it discards---rising to $0.210$\,m at $0.25$\,m, whereas the full representation recovers perfectly at zero and stays at most $0.083$\,m, confirming that $\mathbf{r}_{i,t}$ absorbs mesh-to-point misalignment and decouples binding quality from mesh localisation precision.

\section{Conclusion}
\label{sec:conclusion}
We have presented \emph{MeshBind}, a mesh-guided annotation method that converts articulated mesh deformation into per-point non-rigid scene-flow ground truth. Its core technique employs a per-triangle tangent-normal local frame, which remains invariant under rigid motion yet adapts to articulation. This method is implemented in our developed \emph{SF-Studio}, an interactive tool that incorporates human mesh corrections through a recomputation loop. The SF-Studio offers \emph{SFB} (SUScape-Flow Benchmark), a whole-scene LiDAR scene-flow benchmark that provides per-point non-rigid pedestrian annotations and Rigid / Semi-Rigid / Non-Rigid flow evaluation protocols. Training nine estimators on SFB, either fine-tuned from Argoverse weights or trained from scratch, reduces the three-way EPE by a factor of 2--3$\times$ and brings Non-Rigid EPE to the same level as Rigid EPE. These results show that the labels transfer effectively to conventional supervision and are sufficient as a primary training source. Code, data, and the SF-Studio toolchain will be released upon acceptance.

\section*{Acknowledgments}
DeepSeek assisted with limited drafting and grammar edits; the authors reviewed all text and take full responsibility for the content.

\bibliographystyle{IEEEtran}
\bibliography{ref/refs}

\end{document}